\documentclass[10pt,letterpaper]{article}
\usepackage{tikz}
\usepackage{soul}
\usepackage[pagenumbers ]{cvpr} 
\usepackage{svg}
\usepackage{tabularray}
\usepackage{multirow}
\usepackage{makecell}

\usepackage[toc,page]{appendix}

\usepackage{natbib}
\definecolor{cvprblue}{rgb}{0.21,0.49,0.74}
\usepackage[pagebackref,breaklinks,colorlinks,allcolors=cvprblue]{hyperref}

\title{Multi Anatomy X-Ray Foundation Model}
\author{Nishank Singla\thanks{These authors contributed equally to this work.}, Krisztian Koos\footnotemark[1], Farzin Haddadpour, \\
Amin Honarmandi Shandiz, Lovish Chum, Xiaojian Xu, Qing Jin, Erhan Bas \\
GE HealthCare\\
{\tt\small nishank.singla@gehealthcare.com, erhan.bas@gehealthcare.com}
}
\date{September 2025}

\begin{document}

\maketitle

\begin{abstract}
X-ray imaging is a ubiquitous in radiology, yet most existing AI foundation models are limited to chest anatomy and fail to generalize across broader clinical tasks. In this work, we introduce \textbf{XR-0}, the multi-anatomy X-ray foundation model using self-supervised learning on a large, private dataset of 1.15 million images spanning diverse anatomical regions and evaluated across 12 datasets and 20 downstream tasks, including classification, retrieval, segmentation, localization, visual grounding, and report generation. XR-0 achieves state-of-the-art performance on most multi-anatomy tasks and remains competitive on chest-specific benchmarks. Our results demonstrate that anatomical diversity and supervision are critical for building robust, general-purpose medical vision models, paving the way for scalable and adaptable AI systems in radiology.
\end{abstract}

\section{Introduction}

\begin{figure*}[htp]
    \centering
    \includegraphics[width=0.9\linewidth]{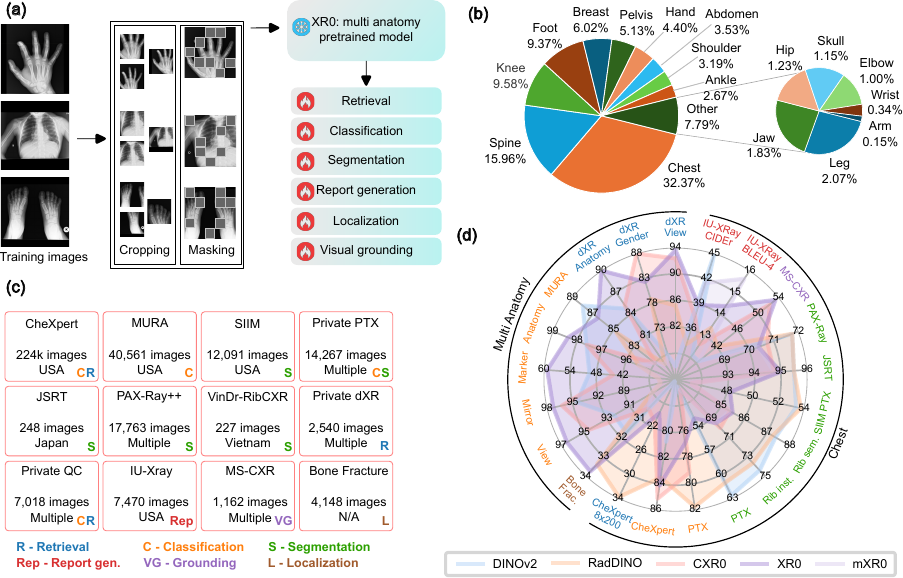}
    \caption{Overview of the XR-0 multi-anatomy pretrained model.
    \textbf{(a)} The model is pretrained using both image-level and patch-level objectives. It is evaluated on a wide range of downstream tasks, including image retrieval, classification, segmentation, report generation, and visual grounding.
    \textbf{(b)} Distribution of anatomical regions in the pretraining dataset. A total of 1.15 million images are used after filtering duplicates and low-quality samples.
    \textbf{(c)} Twelve datasets are used to benchmark model performance across diverse downstream tasks.
    \textbf{(d)} Result summary. XR-0 demonstrate an average convergence speedup of 4.9× on CheXpert, PTX, and QC linear classification tasks compared to DINOv2.}
    \label{fig:overview}
\end{figure*}

Self-supervised learning~(SSL) have demonstrated remarkable capabilities across a wide range of vision and language tasks, particularly when trained on large-scale, diverse datasets~\citep{oquab2023dinov2, he2022masked, chen2020simple, chen2021empirical}, especially when multimodal data is available~\citep{zhang2022contrastive, wang2022medclip, boecking2022making}. In medical imaging, where annotated data is often scarce and costly to obtain, SSL offers a promising avenue for leveraging vast amounts of unlabeled data to learn transferable representations. X-ray imaging, as one of the most widely used diagnostic modalities, presents a compelling domain for developing such models. While public datasets have become increasingly available~\citep{kelly2019key}, they are often limited to specific anatomies—most notably the chest—and lack the diversity needed to support general-purpose medical vision models. Although the quantity and quality of public datasets have improved over the past decade, enabling deep learning models to detect common diseases~\citep{rajpurkar2018deep, rajpurkar2022ai}, annotations and textual reports remain expensive to generate. This has shifted the focus toward developing medical large vision-language models~(LVLMs)~\citep{wolf2023self}. The integration of vision models with large language models (LLMs) has expanded the capabilities of AI systems, enabling tasks such as zero-shot image-to-text generation and natural language-guided image interpretation~\citep{chang2024survey,mu2025mmxu,cho2025perceptionlm, dai2023instructblip}.

In this work, we introduce a multi-anatomy X-Ray pretrained model trained on large and diverse dataset. Our models are based on the vision transformer~(ViT-B) architecture, following DINOv2 SSL framework. The core model, \textbf{XR-0} (Fig.~\ref{fig:overview}a), is trained on 1.15 million X-ray images spanning diverse anatomical regions from head to toe. Additionally, we train \textbf{CXR-0}, a chest-specific model, on the chest subset of the dataset.

We conduct extensive evaluations across 12 datasets and 20 downstream tasks, including image-to-image retrieval, classification, dense segmentation, localization, report generation, and visual grounding. While most public benchmarks focus on chest pathologies, we introduce new multi-anatomy tasks to assess generalization across broader anatomical domains. Our results show that XR-0 achieve state-of-the-art performance on multi-anatomy tasks and remain competitive on chest-specific benchmarks (Fig.~\ref{fig:overview}d), highlighting anatomical diversity and supervision are critical for building robust, general-purpose medical vision models, paving the way for scalable and adaptable AI systems in radiology.

Details about the pretraining and evaluations are provided in Sec.~\ref{sec:methods}, and results are discussed in Sec.~\ref{sec:results}.

\subsection{Related Work}

The development of medical vision foundation models has accelerated in recent years, driven by advances in self-supervised learning (SSL) and the growing availability of public X-ray datasets~\citep{zhou2022generalized, tiu2022expert, zhang2023knowledge}. These models aim to overcome the limitations of supervised learning in medical imaging, particularly the scarcity of labeled data and the narrow anatomical focus of existing datasets.

Early efforts in self-supervised learning (SSL) for medical imaging have predominantly focused on chest X-rays, leveraging either purely visual signals or paired image-text data. Vision-only SSL approaches such as MoCo-CXR~\citep{Sowrirajan2021MoCoCXR} and SimCLR~\citep{Imagawa2024SimCLR} demonstrated the potential of contrastive learning in extracting meaningful representations from unlabeled radiographs. The shift toward multimodal SSL began with ConVIRT~\citep{Zhang2022ConVIRT}, which aligned image and text embeddings using paired radiology reports. Building on this, ELIXR~\citep{xu2023elixr} introduced a two-phase training strategy that aligns a vision encoder with the frozen large language model PaLM 2~\citep{anil2023palm}, further advancing cross-modal representation learning. 

In contrast, other models have explored purely visual pretraining. RadDINO~\citep{perez2025exploring} builds on DINO-based encoder trained on chest X-rays without requiring paired text, addressing the challenge of limited report availability. Similarly, RayDINO~\citep{moutakanni2024advancing} uses self-supervised learning on public chest X-ray datasets to train a large vision transformer. RayDINO is evaluated across 11 datasets and five task categories, and includes a detailed analysis of demographic biases, highlighting the importance of fairness in medical AI.

These studies demonstrate that frozen vision encoders, when paired with lightweight task-specific adapters or heads, can achieve state-of-the-art (SOTA) or near-SOTA performance across a range of tasks. They also underscore the growing emphasis on generalizability and ethical considerations in medical foundation models~\citep{glocker2023bias, li2025embeddings}.

However, most existing models are constrained in two key ways: they are limited to chest imaging, and they rely on a combination of proprietary and public datasets or pretrained weights from general-domain models. In contrast, our work introduces a new class of multi-anatomy X-ray foundation models trained from scratch on a large XR dataset from 12 different clinical sites; across 4 countries in 3 regions.

\section{Experimental Setup (Methods)}
\label{sec:methods}

\subsection{Datasets}

The pretraining dataset is private and consists of 1.17 million X-ray images covering a wide range of anatomical regions from head to toe. The most common anatomies include the chest, spine, knee, foot, breast, and pelvis (Fig.~\ref{fig:overview}b). After filtering duplicates and low-quality images, 1.15 million images are used for pretraining.

For evaluation, we use a combination of public and private datasets spanning multiple tasks and anatomical regions (Fig.~\ref{fig:overview}c). These datasets support tasks such as classification, retrieval, segmentation, localization, and report generation.

\textbf{CheXpert (public):}
A large-scale chest X-ray dataset~\citep{irvin2019chexpert} containing 224,316 images from 65,240 patients. We evaluate disease classification performance using five labels: Cardiomegaly, Edema, Consolidation, Atelectasis, and Pleural Effusion, following the ConVIRT setup~\citep{zhang2022contrastive}.

\textbf{CheXpert 8x200 (public):}
A subset of CheXpert used for image retrieval~\citep{zhang2022contrastive}. It includes 10 query images for each of 8 disease categories, with 200 candidate images per category. The task is to retrieve images with the same label as the query.

\textbf{dXR Retrieval (private):}
A subset of our pretraining dataset used for retrieval tasks based on view (AP/PA/LAT), gender (binary), and image rotation (quadrants). The view retrieval set includes 60 queries and 1,200 candidates; the gender task uses 80 queries and 1,200 candidates.

\textbf{Pneumothorax (PTX) (private):}
This dataset comprises 11,997 frontal chest X-ray images labeled for pneumothorax presence, including 5,019 positive and 6,978 negative cases. The validation set contains 1,846 cases, of which 835 are positive, while the test set includes 424 cases with 177 positives. Additionally, pixel-level segmentations are provided to support segmentation tasks. 

\textbf{Quality Control (QC) (private):}
Contains 7,018 images labeled for marker presence (binary), visible anatomy (13 classes), projection view (6 classes), and mirrored state (5 classes). 
The full list of supported classes is presented in Appendix \ref{apdx:qc_classes}. 
A subset is used for quadrant retrieval (40 queries, 640 candidates labeled by 90° rotation).

\textbf{MURA (public):}
A dataset of 40,561 multi-view upper extremity radiographs~\citep{rajpurkar2017mura}, labeled as normal or abnormal by radiologists. Predictions are aggregated per study, following the original evaluation protocol.
  
\textbf{VinDr-RibCXR (public):}
Provides segmentation masks for all 20 ribs across 196 training and 49 validation images~\citep{nguyen2021vindr-rib}. The official validation set is used for testing.

\textbf{SIIM-ACR PTX (public):}
A pneumothorax segmentation dataset~\citep{siim-acr-pneumothorax-segmentation} with over 12,000 samples. We use the stage 2 validation set for testing.

\textbf{JSRT (public):}
Includes 247 chest X-rays with annotations for lungs, clavicles, and heart~\citep{shiraishi2000jsrt}. The dataset contains 154 nodule and 93 non-nodule cases and is widely used for segmentation and localization tasks.

\textbf{PAX-Ray++ (public):}
A comprehensive segmentation dataset with 157 anatomical classes (soft tissues and bones) across 7,377 images~\citep{Seibold_2022_BMVC, seibold2023accurate}. 

\textbf{IU-XRay (public):}
A multimodal dataset with 7,470 image–report pairs~\citep{demner2016preparing}, commonly used for report generation. We generate the `Findings' and `Impression' sections from the input image.

\textbf{MS-CXR (public):}
Contains 1,162 image–sentence pairs with bounding boxes and corresponding descriptive phrases for eight radiological findings~\citep{boecking2022making, goldberger2000physiobank}. Used for visual grounding tasks.

\textbf{Bone Fracture Detection (public):}
The dataset includes bounding box annotations for 4,148 images, which are divided into 3,631 for training, 348 for validation, and 169 for testing~\citep{darabi2024bone}. The annotated images cover a range of anatomies, including the elbow, fingers, forearm, humerus, shoulder, and wrist.


\subsection{Pretraining}

\subsubsection{Data Filtering}

All DICOM images are converted to 8-bit unsigned integer format and saved as PNG files. Each image is resized such that the shortest side is 1024 pixels, while maintaining the original aspect ratio. To identify and remove duplicate, noisy, or empty scans, we compute image embeddings using the CLIP ViT-L model~\citep{radford2021learning}. These embeddings are then used for near-duplicate detection and image similarity search to filter out redundant or low-quality images.
After filtering, 1.15 million high-quality images remain for model pretraining.

\subsubsection{Training Setup}

We adopt the DINOv2 architecture~\citep{oquab2023dinov2}, where the training objective combines multiple components:
\begin{itemize}
    \item DINO loss for image-level representation learning, using both global and local crops.
    \item KoLeo loss as a regularization term to encourage uniformity in the embedding space.
    \item iBOT loss for patch-level learning via masked image modeling.
\end{itemize}

Input images are zero-padded symmetrically to center the content, normalized using ImageNet mean and standard deviation, and resized to 518×518 pixels. Global crops cover 50–100\% of the original image and are resized to 518×518, while local crops cover 20–50\% and are resized to 196×196. In an ablation study, we also resized images to 518 pixels while preserving aspect ratio, and observed no significant difference during pretraining. 
Training is performed on a single node with 8 NVIDIA A10G GPUs (24GB each). We use PyTorch’s Fully Sharded Data Parallel (FSDP) with mixed precision (FP16). The batch size is 10 per GPU, resulting in a global batch size of 80. Optimization is done using AdamW with a learning rate of 1e-3, weight decay ranging from 0.04 to 0.2, and a cosine annealing learning rate schedule with warm restarts.
To monitor progress during pretraining, we include a lightweight evaluation task based on CheXpert image retrieval (see Sec.~\ref{sec:retrieval-tasks}). This task involves retrieving images for 8 disease categories, each with 200 candidates.

We train two primary vision models: \textbf{XR-0} -- A multi-anatomy model trained on the full dataset, covering diverse anatomical regions; and \textbf{CXR-0} -- A chest-specific model trained on a subset containing only chest X-rays.

\subsection{Evaluation tasks}

We evaluate our models across a diverse set of tasks, covering retrieval, classification, segmentation, localization, visual grounding, and report generation. Unless otherwise specified, all experiments are conducted on a single NVIDIA A10 GPU with 24GB memory. Comparison tables show the best scores in bold and the second-best is textitd.

\subsubsection{Image-Retrieval}
\label{sec:retrieval-tasks}

The goal is to retrieve semantically similar images for a given query based on embedding similarity. We use the \texttt{[CLS]} token output from the vision transformer as the image embedding and compute cosine similarity between query and candidate embeddings.

Each query image is used to retrieve the top-$k$ most similar images from a candidate set. The evaluation metric is Precision@$k$, computed based on image-level labels (e.g., presence of the same disease in both query and candidate images).

\subsubsection{Classification}
\label{sec:classification-tasks}

We evaluate binary, multi-class, and multi-label classification tasks using a frozen backbone and a 3-layer MLP classifier with ReLU activations and dropout ($p = 0.2$). We use cross-entropy~(CE) loss for multi-class tasks, and binary-CE for multi-label scenarios. The classifier is trained on precomputed \texttt{[CLS]} embeddings (i.e. the backbone is frozen). The default setup for training the models is batch size of 64, maximum 500 epochs, Adam optimizer, base learning rate of 1e-4, that is reduced by a factor of 2 when reaching a plateau, LR patience is 2, weight decay of 1e-6, and early stopping if the validation score does not improve for 10 straight epochs.

The AUROC or Matthews Correlation Coefficient~(MCC) scores are reported on the corresponding test set. To measure the efficiency of the models, we train classification tasks at different data regimes, such as 1\%, 10\%, and 100\% of the training data when there is enough data support for the percentage split. The limited data experiments are repeated five times (with different random seeds), ensuring that different models are always trained on the same splits, and the average score is reported.

\subsubsection{Segmentation}

We evaluate segmentation performance on public and private datasets using two decoder designs: a linear head (single convolution with upsampling) and UPerNet~\citep{xiao2018unified}, with the backbone frozen in all cases. The linear decoder evaluates the representational quality of patch embeddings, while UPerNet leverages multi-scale features for enhanced segmentation. We compare performance against two baselines: the general-domain DINOv2 and the chest-specialized RadDINO.

Models are trained for 5,000 iterations with early stopping, using the Dice Similarity Coefficient (DSC) as both the training objective and evaluation metric. Training uses a batch size of 8 and a base learning rate of 1e-4, reduced by a factor of 2 upon validation DSC plateau. Data augmentations include random affine (p=0.5; translate=40; rotate=$\pi$/36; scale=0.15), elastic transforms (p=0.1; spacing=200; magnitude=(0,2); rotate=$\pi$/72; shear=0.1; translate=10; scale=(0.05, 0.15)), gamma adjustment (p=0.1; range=(0.9, 1.2)), and Gaussian noise (p=0.1; mean=0; std=0.05).

\subsubsection{Localization and Visual Grounding}
\label{sec:localization-tasks}

We evaluate multi-anatomy localization performance on the publicly available Bone Fracture Detection dataset~\citep{darabi2024bone}, which comprises lower and upper extremities X-ray images annotated for fracture detection. To this end, we train a DETR-based decoder~\citep{carion2020end} for 90 epochs using a combination of $\mathcal{L}_1$ and Intersection-over-Union (IoU) loss functions. Training is conducted with a batch size of 4 and a fixed learning rate of 1e-5 where we freeze the encoder and only update decoder weights. We report mean Intersection over Union (mIoU) and mean Average Precision at an IoU threshold of 0.5 (mAP@50).

We test the grounding capability of models following the setup described in~\citep{chen2023medrpg}, replacing the vision encoder with pretrained models and attaching localization heads. Accuracy (Accu $>$ 50\%) and mean Intersection over Union (mIoU) are used as evaluation metrics on MS-CXR benchmark~\citep{boecking2022making}. 

We use BERT~\citep{devlin2019bert} as language encoder and 6 block cross attention block for multimodal fusion between vision and language encoders following~\citep{dou2022empirical}. For the ResNet-50 baseline, we use random initialization and train the model end-to-end, as the original weights were not publicly available. For other models, we use linear probing setup where only the mapping layer is tuned. Similarly, limited annotation experiments were carried out for 10\% and 100\% data regimes. 

\subsubsection{Report Generation}

We evaluate report generation using the IU-XRay dataset~\citep{demner2016preparing}, following the prefix tuning setup~\citep{wang2023r2gengpt, mokady2021clipcap, wang2022ofa}. A linear projection (MLP) connects the frozen vision encoder to a frozen language decoder (LLaMA-7B-Chat~\citep{nousresearch_llama2_7b_chat}). All models are trained with 8 NVidia A100 80GB GPUs with a local batch size of 2 for visual encoders.

We select the best checkpoint based on the average of BLEU-4 and CIDEr scores on the validation set. Final evaluation includes BLEU-1 to BLEU-4, ROUGE-L, and CIDEr~\citep{wang2023r2gengpt, hyland2023maira, saab2024capabilities}.

To assess annotation efficiency, we repeat training with 10\% of the IU-XRay dataset and compare results to the full-data setup, following the same validation protocol as in classification and localization tasks (Sec.~\ref{sec:classification-tasks}, \ref{sec:localization-tasks}).

\subsection{Comparing Pretrained Approaches}

We compare our models against DINOv2 and RadDINO. DINOv2 is pretrained on the LVD-142M dataset of natural images~\citep{baharoon2024evaluatinggeneralpurposevision, tayebi2024enhancing}, representing a general-domain SSL baseline.
RadDINO is a chest X-ray–specialized model trained on five public datasets totaling 882,775 chest images~\citep{perez2025exploring}, considered state-of-the-art (SOTA) for chest imaging tasks.
All models use the Vision Transformer Base (ViT-B) architecture with a patch size of 14, comprising approximately 86 million parameters. While DINOv2 is not trained on medical data, it has shown strong generalization in prior work~\citep{baharoon2023evaluating,muller2025medical}. RadDINO, on the other hand, is optimized for chest-specific tasks.

We evaluate all models across six task categories: image retrieval, classification, segmentation, localization, visual grounding, and report generation, as described in Sec.~\ref{sec:methods}

\section{Results}
\label{sec:results}

\subsection{Anatomy-Aware Performance Trends in Pretrained Models}

\subsubsection{Performance in Image Retrieval}
\begin{figure}[tbp]
    \centering
    \includegraphics[width=1\columnwidth]{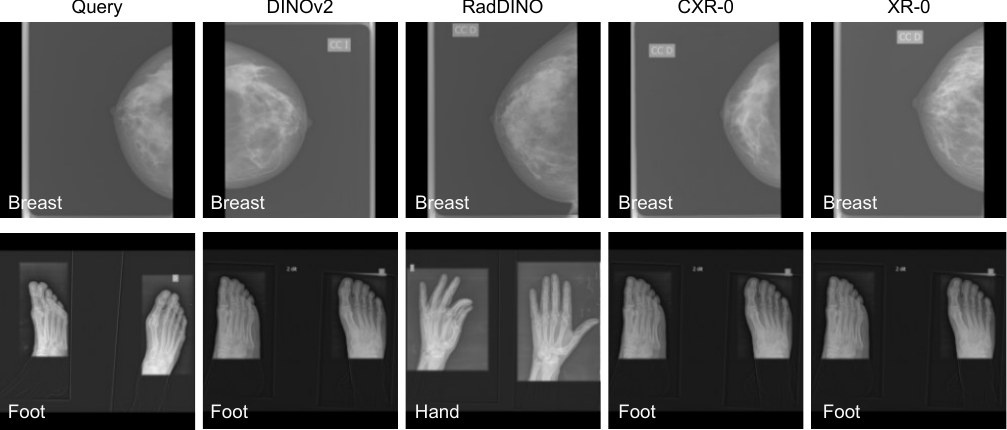}
    \caption{Image retrieval examples from the dXR Anatomy dataset. While all models retrieve relevant images based on the target class, domain-specific models often return results that are more consistent in secondary visual attributes such as texture, orientation, or anatomical presentation.}
    \label{fig:retrieval_results}
\end{figure}

We extensively evaluate the retrieval capabilities of our models on both public and private datasets. The public CheXpert 8x200 dataset~\citep{zhang2022contrastive} is used for disease-based retrieval, while proprietary dXR datasets are used for tasks such as view estimation, gender classification, anatomy identification, and quality control (Tab.~\ref{tab:retrieval}). Example results are shown in Fig. \ref{fig:retrieval_results} and Appendix \ref{apdx:retrieval_examples}. 

All X-ray–specific pretrained models outperform the general-domain DINOv2 baseline, underscoring the importance of in-domain pretraining for medical image retrieval. Among chest-specific models, RadDINO achieves the highest performance on the CheXpert retrieval task (P@5: 33.8), likely due to its use of CheXpert data during pretraining. However, CXR-0, which does not use CheXpert during training, outperforms RadDINO on the dXR View retrieval task (P@5: 92.7 vs. 87.0), suggesting that task-specific data diversity can compensate for dataset overlap.

Interestingly, XR-0, the multi-anatomy model, achieves the best performance on generic tasks such as view and anatomy retrieval—even when the anatomy is chest-related. This indicates that exposure to a broader range of anatomical contexts enhances the model's generalization ability. For example, XR-0 achieves the highest P@5 score (93.7) on dXR View retrieval and performs competitively across other tasks.

These findings highlight several key insights. 
In-domain pretraining significantly boosts retrieval performance, especially for specialized tasks. 
Multi-anatomy pretraining improves generalization, making models more robust across diverse clinical scenarios. 
Data diversity and task alignment are critical for optimizing retrieval performance in real-world healthcare applications.

From a clinical perspective, improved retrieval models can enhance decision support systems by surfacing relevant prior cases, aiding in diagnosis, and reducing diagnostic variability. This is particularly valuable in settings with limited expert availability or high case volumes.

\begin{table*}[tbp]
\centering
    \begin{tabular}{l ccc ccc ccc ccc}
    \toprule
 & \multicolumn{3}{c}{CheXpert}& \multicolumn{3}{c}{dXR View} & \multicolumn{3}{c}{dXR Gender} & \multicolumn{3}{c}{dXR Anatomy}\\
 
         Models&  @5&  @10&  @100&  @5& @10& @100& @5& @10&@100 & @5& @10&@100\\
         \hline
         DINOv2
&  18.2&  19.0&  15.7&  79.7& 78.7& 70.7& 71.8& 69.8&60.4 & 85.8& \textit{82.7} & \textbf{61.5}\\
         RadDINO
& \textbf{33.8} & \textbf{31.9} & \textbf{22.3} & 87.0 & 83.7 & 76.0 & 80.8 & 79.6 & \textit{63.4} & 79.7 & 75.9 & 48.6\\
        \hline
      CXR-0&  \textit{23.8}&  \textit{23.6}&  \textit{18.5}&  \textit{92.7}& \textit{89.5}& \textbf{78.5}& \textbf{88.0}& \textbf{83.5}&62.7 & \textit{86.2} & 81.7 & 52.5\\
     XR-0 &  22.5 &  \textit{23.6} &  18.2 &  \textbf{93.7} & \textbf{90.0} & \textit{77.9} & \textit{83.5} & \textit{82.0} & \textbf{66.4} & \textbf{90.0} & \textbf{86.7} & \textit{59.6}\\
    \bottomrule
    \end{tabular}
    \caption{Precision@$k$ measures the fraction of relevant images among the top-$k$ retrieved results in image retrieval tasks.}
    \label{tab:retrieval}
\end{table*}

\subsubsection{Performance in Classification}
\label{sec:results-classification}
Classification, supporting automated diagnosis of diseases and abnormalities, is ubiquitous in X-ray imaging. We benchmark model performance on public and internal datasets, including CheXpert, PTX, MURA and QC.

Chest-specific models such as RadDINO perform best on chest-related tasks, as expected. However, our models consistently outperform the general-domain DINOv2 baseline and remain competitive across most settings (Tab.~\ref{tab:classification}). Notably, CXR-0 achieves the highest AUROC on CheXpert with full data (85.3), while RadDINO leads on PTX (81.8). 

Surprisingly, DINOv2 achieves the best performance on the MURA dataset, which consists of musculoskeletal radiographs. This may be due to its large-scale pretraining on 140M natural images, compared to 1.15M X-rays for XR-0. 

XR-0 demonstrates strong performance on internal hold-out QC classification tasks (Tab.~\ref{tab:classification-QC}), outperforming other models in general. Specifically, XR-0 achieves a gain of +0.6 MCC in anatomy classification over the second best model, +7.7 MCC in marker detection, +1.2 MCC in mirror state prediction, and +0.8 MCC in view classification when leveraging all available data. Such gain is more pronounced when limited annotation experiments are caried out at lower data fractions. We observe baseline DINOv2 model consistently outperforms chest specialized SOTA RadDINO model on multi-anatomy classification tasks, which further emphasize the importance of data/anatomy diversity in building XR pretrained models. 

These results highlight the value of multi-anatomy pretraining for general-purpose medical imaging tasks. In clinical settings, such models can support a wide range of diagnostic workflows, from disease detection to quality control.

Our findings suggest that the \texttt{[CLS]} token may inadequately capture task-specific information. For instance, lead marker classification using \texttt{[CLS]} embeddings yields suboptimal performance. In contrast, patch embeddings reveal preserved spatial features relevant to lead markers. PCA visualization of the patch embeddings is shown in the Appendix~\ref{apdx:pca_of_patchemb}.

Replacing the \texttt{[CLS]} token with patch embeddings improves the performance by 25.0 MCC score~(absolute) for XR-0, underscoring the effectiveness of spatially distributed representations over global summarization.

\begin{table*}[htbp]
    \centering
    \begin{tabular}{l ccc ccc ccc}
    \toprule
 & \multicolumn{3}{c}{CheXpert (AUC)}& \multicolumn{3}{c}{PTX (AUC)} 
 & \multicolumn{3}{c}{MURA (AUC)}\\
         Models&  1\%&  10\%&  all&  1\%&  10\%& all&  1\%& 10\%& all 
    \\
    \hline
         DINOv2 &  78.3 &  81.2 &  81.8 &  65.5 & 71.7 & 74.4 & \textbf{73.8} & \textbf{84.2} & \textbf{88.2} \\
         RadDINO &  78.4 &  82.7 &  84.0 &  \textbf{71.4} & \textbf{78.3} & \textbf{81.8} & 71.5 & 81.3 & 85.6
    \\
    \hline
         CXR-0 &  \textit{79.0} &  \textbf{83.6} &  \textbf{85.3} &  \textit{68.3} & \textit{74.9} & \textit{79.7} & 66.3  & 76.8 & 82.6
    \\
         XR-0 &  \textbf{80.2} &  \textit{83.3} &  \textit{84.5} &  67.7 & 74.6 & 77.9 & \textit{72.3} & \textit{82.8} & \textit{87.0}

    \\
    \bottomrule
    \end{tabular}
    \caption{Performance on CheXpert, PTX, and MURA classification tasks and training set fractions (AUROC).}
    \label{tab:classification}
\end{table*}

\begin{table*}[htbp]
    \centering
    \begin{tabular}{l ccc ccc cc ccc}
    \toprule
 & \multicolumn{3}{c}{Anatomy (MCC)}& \multicolumn{3}{c}{Marker (MCC)} & \multicolumn{2}{c}{Mirror (MCC)} & \multicolumn{3}{c}{View (MCC)}\\

         Models&  1\%&  10\%&  all&  1\%&  10\%& all& 
         10\%& all&  1\%& 10\%& all\\
         \hline
         DINOv2
& \textit{76.9} & \textit{93.9} & \textit{97.9} & \textbf{8.4} & \textit{36.8} & \textit{51.7} & 
71.6 & 95.2 & \textit{64.3} & 86.4 & 93.3
\\
         RadDINO
& 65.6 & 89.7 & 96.2 & 2.2 & 23.6 & 39.1 & 
57.9 & 87.4 & 53.7 & 81.5 & 90.7
\\
\hline
         CXR-0
& 74.5 & 93.6 & 97.9 & 5.7 & 28.3 & 47.4 & 
\textit{92.7} & \textit{96.6} & 55.7 & \textit{87.3} & \textit{95.1}
\\
         XR-0
& \textbf{85.7} & \textbf{96.5} & \textbf{98.5} & \textit{6.7} & \textbf{36.8} & \textbf{59.4} & 
\textbf{93.6} & \textbf{97.4} & \textbf{72.7} & \textbf{93.1} & \textbf{96.6}
\\
    \bottomrule
    \end{tabular}
    \caption{Performance on QC classification tasks. (Note, that Mirror has no 1\% scores due to lack of training data in this setting.}
    \label{tab:classification-QC}
\end{table*}

\subsection{Architectural Trade-offs in Dense Visual Understanding}
\subsubsection{Performance in Chest Region Segmentation}
\label{sec:results-segmentation}
We evaluate segmentation performance across six tasks spanning five datasets, all focused on chest anatomy (Tab.~\ref{tab:segmentation}). 
Using a linear decoder, all X-ray–specific models outperform the general-domain DINOv2 baseline. Notably, XR-0 achieves the best performance on the JSRT dataset and ranks second across all other tasks. This strong performance is particularly noteworthy given that XR-0 is trained entirely from scratch—without warm-start initialization—and uses a relatively modest dataset (~380K images, including ~220K frontal chest X-rays), in contrast to chest-specific models like RadDINO. These results highlight the effectiveness of multi-anatomy pretraining in enabling robust generalization across anatomy-specific tasks.

\begin{figure}[tbp]
    \centering
    \begin{tabular}{@{}c@{}c@{}c@{}c}
        & \textbf{Original} & \textbf{Prediction(XR0)} & \textbf{Ground Truth} \\

        \rotatebox{90}{\hspace{0.4cm}SIIM PTX} &
        \includegraphics[width=0.25\columnwidth]{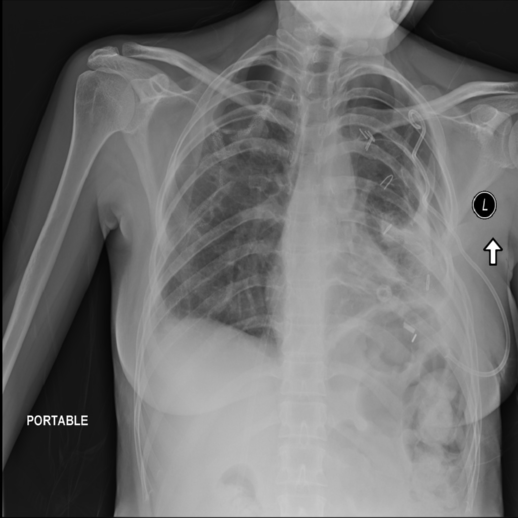} &
        \includegraphics[width=0.25\columnwidth]{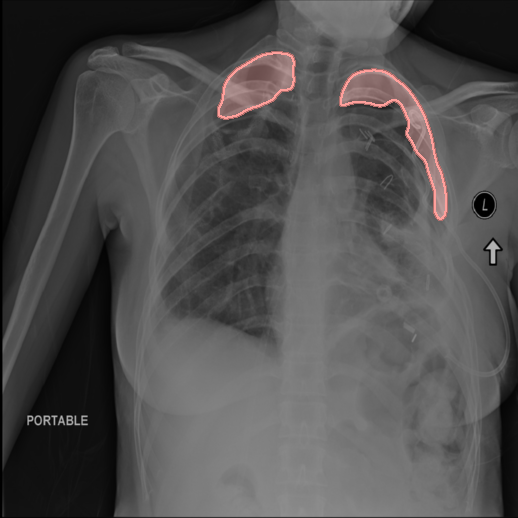} &
        \includegraphics[width=0.25\columnwidth]{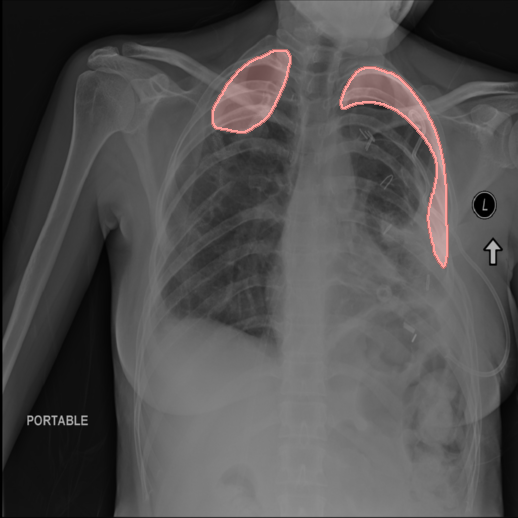}  \\
        
        \rotatebox{90}{\hspace{0.4cm}Rib instance} &
        \includegraphics[width=0.25\columnwidth]{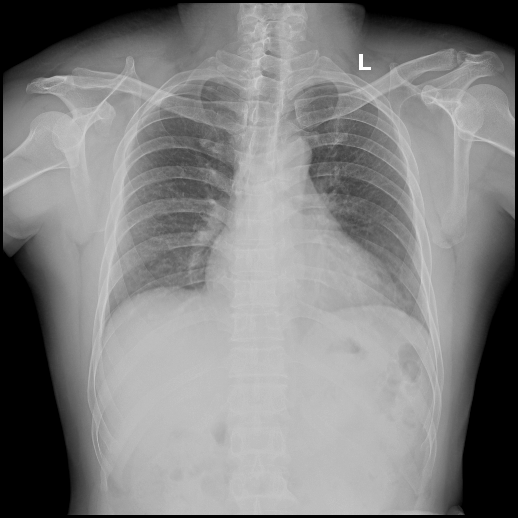} &
        \includegraphics[width=0.25\columnwidth]{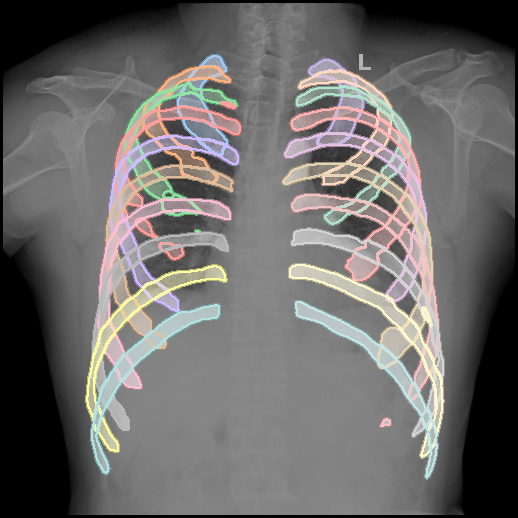} &
        \includegraphics[width=0.25\columnwidth]{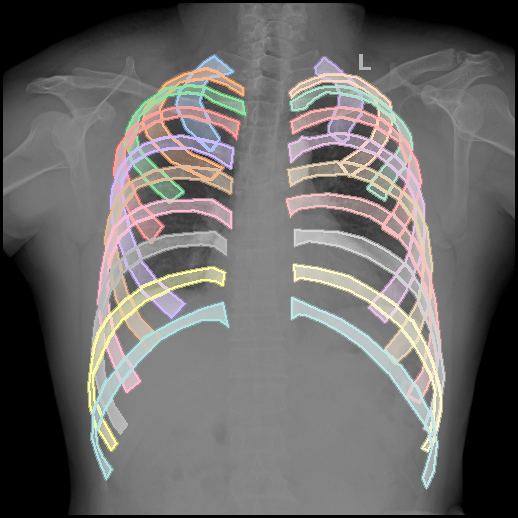} \\
        
        \rotatebox{90}{\hspace{0.4cm}JSRT} &
        \includegraphics[width=0.25\columnwidth]{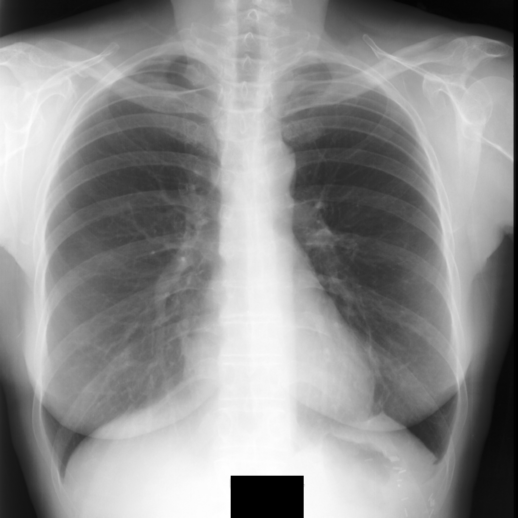} &
        \includegraphics[width=0.25\columnwidth]{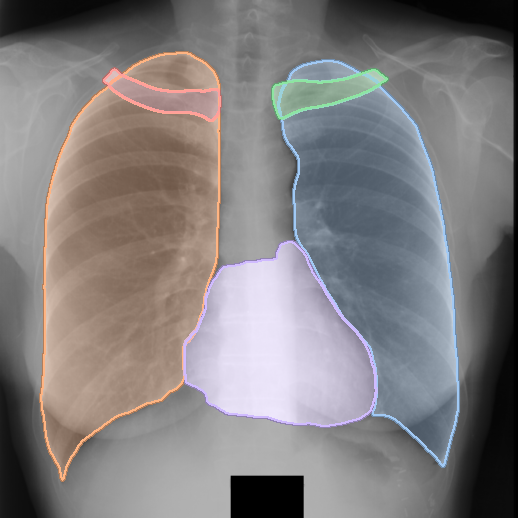} &
        \includegraphics[width=0.25\columnwidth]{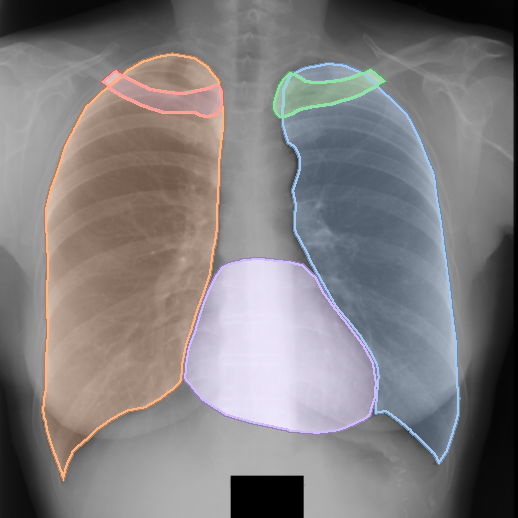} \\ 
        
        \rotatebox{90}{\hspace{0.4cm}PAX-Ray++} &
        \includegraphics[width=0.25\columnwidth]{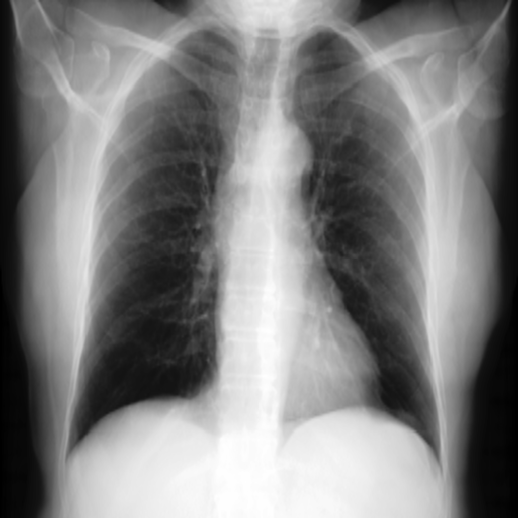} &
        \includegraphics[width=0.25\columnwidth]{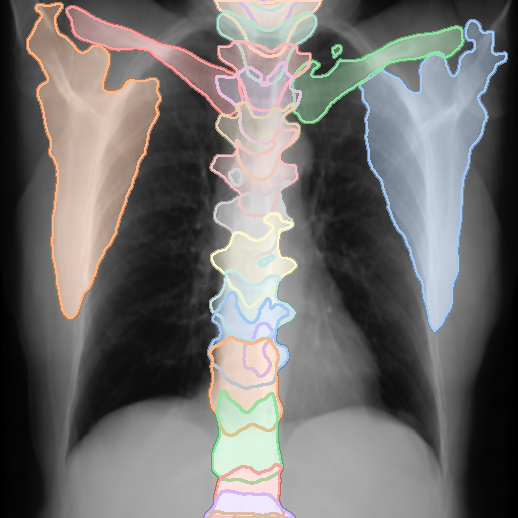} &
        \includegraphics[width=0.25\columnwidth]{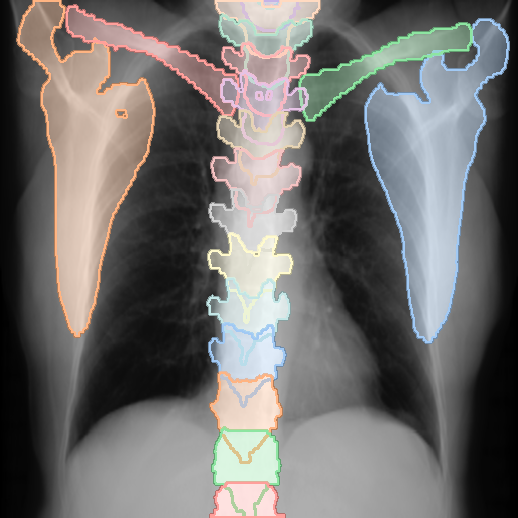} \\
    \end{tabular}
    \caption{Qualitative segmentation results using the XR-0 model with a UPerNet decoder. Only the decoder is trained; the backbone remains frozen.}
    \label{fig:segmentation_results}
\end{figure}

However, when evaluated with the UPerNet decoder (Fig. \ref{fig:segmentation_results}, CXR-0 and XR-0 under-perform relative to DINOv2 and RadDINO. We hypothesize that this discrepancy stems from differences in intermediate feature quality. DINOv2, pretrained on large-scale natural images, consistently delivers top segmentation results, while RadDINO inherits these robust features via warm-start initialization. In contrast, our models are trained from scratch, which may limit the expressiveness of intermediate representations required by complex decoders like UPerNet.

The ability of XR-0 to perform competitively with a simple linear decoder highlights its potential for deployment in resource-constrained settings. The observed architectural trade-offs further emphasize the need to align decoder complexity with the quality of learned representations for optimal performance.

\begin{table*}[htbp]
    \centering
    \begin{tabular}{l c ccc ccc ccc}
    \toprule
 & & PTX & VinDr-RibCXR & VinDr-RibCXR  & SIIM-ACR & JSRT & PAX-Ray++\\
         Models & Decoder & private & instance & semantic & PTX & & \\
         \hline
         DINOv2 & Linear &
         30.8 &	38.7 &	73.1 &	27.8 &	82.8 &	50.0 \\
         RadDINO & Linear &
         \textbf{38.6} & \textbf{49.9} & \textbf{73.8} & \textbf{37.0} & \textit{86.1} & \textbf{59.3} \\
         CXR-0 & Linear &
         32.9 & 41.3 & 73.3 & 29.1 & 82.6 & 55.3 \\
         XR-0 & Linear &
         \textit{34.0} & \textit{42.7} & \textit{73.6} & \textit{32.6} & \textbf{86.2} & \textit{56.3} \\
         \hline
         DINOv2 & UPerNet &
         \textbf{62.7} & \textbf{74.5} & \textbf{87.5} & \textit{53.5} & \textit{95.6} & \textit{71.8} \\
         RadDINO & UPerNet &
         \textit{61.4} & \textit{73.9} & \textit{87.4} & \textbf{53.8} & \textbf{95.6} & \textbf{71.8} \\
         CXR-0 & UPerNet &
         52.0 & 71.0 & 85.9 & 47.0 & 95.0 & 71.0 \\
         XR-0 & UPerNet &
         54.9 & 70.6 & 86.0 & 49.4 & 95.2 & 70.7 \\
         \bottomrule
    \end{tabular}
    \caption{Segmentation performance (DSC) using frozen backbones with linear and UPerNet decoders.}
    \label{tab:segmentation}
\end{table*}

\subsubsection{A case study on multi-anatomy model on a specialized task}
We select the internal PTX task to further investigate how architectural and training design choices affect downstream performance. This task is particularly relevant in clinical settings, as pneumothorax (PTX) is a potentially life-threatening condition that requires accurate and timely detection.

First, we evaluate the impact of input resolution. During pretraining, models were trained with an input size of $518 \times 518$. For this experiment, we increase the resolution to $1022 \times 1022$ (a multiple of the patch size 14) and train the models end-to-end using Dice and Focal losses. 

Increasing the resolution results in a +7.37 improvement in AUROC score (reaching an absolute value of 88.9) for the XR-0 model, compared to the baseline reported in Table~\ref{tab:ptx-design-choice}.
This highlights the importance of high-resolution inputs for detecting small or subtle findings, such as minor PTX regions.

Next, we explore different architectural designs for leveraging the segmentation masks available in the PTX dataset, using the XR-0 model. While the standard classification setup uses the \texttt{[CLS]} token, this approach does not exploit the spatial information encoded in the segmentation masks. To address this, we evaluate four model variants:

\begin{enumerate}
    \item {MLP on \texttt{[CLS]} token:} A conventional classification setup used throughout this study.
    \item {Convolutions on patch embeddings:} A decoder with three bottleneck-style convolutional layers (1$\times$1, 3$\times$3, 1$\times$1), followed by adaptive pooling and a linear classifier.
    \item {Multitask:} A dual-head model that outputs both a classification score and a segmentation mask. The classification head uses the same convolutional decoder as above; the segmentation head is a linear decoder with upsampling.
    \item {Multitask+:} An enhanced multitask setup with five convolutional layers and a larger hidden dimension (1/2 of the ViT feature size). Dice and structure losses~\citep{fan2020pranet} are used for optimization.
\end{enumerate}

\begin{table}[htbp]
\centering
\begin{tabular}{|c|c|c|c|c|c|}
\hline
 \textbf{XR-0} & Input res. & CLS MLP & PE Conv & Multitask & Multitask+ \\
\hline
AUC & 518 & 81.5 & 83.7 & 87.3 & 88.7 \\
$\Delta$ & & & +2.2 & +3.5 & +1.5\\
\hline
\end{tabular}
\caption{Impact of model and setup design on PTX classification. Using patch embeddings and segmentation-aware multitask learning improves AUC.}
\label{tab:ptx-design-choice}
\end{table}

The results (Tab.~\ref{tab:ptx-design-choice}) show that more complex architectures can improve classification performance. Notably, the multitask+ setup achieves the highest AUC of 88.7, demonstrating the benefit of incorporating spatial supervision and richer decoder designs.

To further improve performance, we increase the input resolution beyond $1022 \times 1022$. As shown in Tab.~\ref{tab:ptx-inputres}, increasing the resolution to $1540 \times 1540$ in multitask+ setup yields additional gains in both classification (AUC) and segmentation (Dice) performance (experiments performed with a reduced batch size of 8 instead of 16 to fit GPU memory for high resolution models). The best results are achieved when XR-0 is trained on a combination of internal and SIIM datasets, reaching an AUC of 95.1 and Dice score of 67.8.

\begin{table}[htbp]
\centering
\begin{tabular}{|c|c||c|c|}
\hline
Model & Input res. & AUC & DicePos \\
\hline
XR-0 & 518 & 88.1 & 59.3 \\
XR-0 & 1022 & 92.5 & 66.5 \\
XR-0 & 1540 & \textit{94.9} & \textit{66.7} \\
XR-0* & 1540 & \textbf{95.1} & \textbf{67.8} \\
\hline
\end{tabular}
\caption{Performance of the multitask+ setup across varying input resolutions. XR-0* is trained on a combination of internal and SIIM datasets.}
\label{tab:ptx-inputres}
\end{table}

These findings emphasize the importance of architectural flexibility and high-resolution inputs for detecting subtle pathologies. In clinical practice, such improvements could lead to earlier and more accurate detection of pneumothorax, potentially reducing diagnostic delays and improving patient outcomes.

\subsubsection{Performance in Localization and Visual Grounding Tasks}

\begin{table}[ht]
  \centering
  \begin{tabular}{lcccc}
        \toprule
        & \multicolumn{2}{c}{Bone Fracture} & \multicolumn{2}{c}{MS-CXR (mIoU)}\\
         \cmidrule(lr){2-3} \cmidrule(lr){4-5}
        Model & mAP50 & mIoU & 10\% & all\\
        \midrule
        DINOv2         & 19.7 & 32.2 & 29.1 & 41.6 \\
        RadDINO      & \textit{21.8} & \textit{33.5} & 30.0 & 44.5\\
        \hline
        CXR-0         & 14.5 & 31.2  & \textit{31.0} & \textit{52.6}\\
        XR-0          & \textbf{26.0} & \textbf{33.8} & \textbf{33.1} & \textbf{53.5}\\
        \bottomrule
    \end{tabular}
    \caption{Bone Fracture Localization and Visual grounding performance}
    \label{tab:localization_and_grounding_results}
\end{table}

We evaluate localization performance on the Bone Fracture Detection dataset using a frozen backbone as shown in Tab.~\ref{tab:localization_and_grounding_results}~(qualitative examples in Appendix~\ref{apdx:localization_examples}). 
Overall the XR-0 model~(mAP@50: 26.0, mIoU: 33.8) outperforms both general-purpose (DINOv2) and chest-specialized (RadDINO) models.

Visual grounding results on the MS-CXR dataset are shown in Tab.~\ref{tab:localization_and_grounding_results}~(qualitative examples in Appendix~\ref{apdx:grounding_examples}). 
XR-0 achieves the best performance across different data annotation regimes~( 10\% and 100\% with mIoU scores of 33.1 and 53.5, respectively). Compared to DINOv2 and RadDINO, XR-0 shows a relative improvement of 28.8\% and 20.4\% when all data is used, respectively.

Accurate localization and grounding are essential for tasks such as fracture detection, lesion annotation, and radiology report alignment. The ability of XR-0 to perform well in these tasks, especially under limited supervision, makes it a promising candidate for deployment in real-world diagnostic workflows where labeled data is scarce.

\subsection{Enhanced Report Generation via Multi-Anatomy and Multi-Modal Learning}
We evaluated report generation performance on the IU-XRay dataset using standard NLP metrics, including BLEU-[1-4], ROUGE-L, and CIDEr (Tab.~\ref{tab:repgen}).
We further extended XR-0 into a multimodal framework, termed mXR-0, by integrating 69,000 paired clinical reports along with a dedicated text encoder. For multimodal pretraining, we initialize the vision encoder with the pretrained XR-0 weights and adopt PubMedBERT~\citep{pubmedbert}, a language model pretrained on biomedical literature, as the text encoder. Both encoders are jointly fine-tuned using an image–text contrastive learning objective inspired by CLIP~\citep{clip}, which aligns image and text embeddings in a shared multimodal space by pulling together matched pairs and pushing apart mismatched ones. 
This enhancement enabled us to evaluate the impact of weak supervision from clinical narratives on the quality of generated reports. Our initial findings highlight the benefits of both multi-anatomy and multimodal pretraining for generative tasks in medical imaging.

First, when using the full training dataset, both XR-0 and mXR-0 outperform all other models across most metrics. mXR-0 achieves the highest scores in BLEU-1 to BLEU-4, while XR-0 leads in ROUGE-L when using all data. The only exception is DINOv2, which achieves the highest CIDEr score (44.9), likely due to its large-scale pretraining on natural images. However, its performance on other metrics remains significantly lower.

Second, in the low-data regime (10\% of training data), both XR-0 and mXR-0 outperform RadDINO and DINOv2 across all metrics. This demonstrates the robustness of our visual encoders, which generalize well even with limited supervision. Notably, mXR-0 achieves the best BLEU-1 (39.0), and BLEU-2 (24.9) scores in this setting, highlighting the value of multimodal pretraining.

These results underscore the importance of domain-specific and multimodal pretraining for generative tasks in medical imaging. Automated report generation can reduce radiologist workload, improve reporting consistency, and support decision-making—especially in resource-constrained settings where expert annotations are limited.

\begin{table*}[!htbp]
    \centering
    \begin{tabular}{l cc cc cc cc cc cc cc}
    \toprule
    & \multicolumn{2}{c}{BLEU-1} & \multicolumn{2}{c}{BLEU-2} & \multicolumn{2}{c}{BLEU-3} & \multicolumn{2}{c}{BLEU-4} & \multicolumn{2}{c}{ROUGE-L} & \multicolumn{2}{c}{CIDEr} \\
    Models & 10\% & all & 10\% & all& 10\% & all & 10\% & all & 10\% & all & 10\% & all \\
    \hline
    DINOv2
    & 16.8 & 36.2 & 09.5 & 23.4 & 06.3 & 16.9 & 04.3 & 12.9
    & 13.4 & 34.6
    & 06.3 & \textbf{44.9}\\
    RadDINO
    & 30.9 & \textit{42.5} & 19.7 & \textit{27.2} & 13.7 & 19.0 & 09.8 & 13.9
    & 31.9 & 36.3
    & 18.9 & 36.4\\
    \hline
    CXR-0
    & 36.8 & 36.8 & 23.9 & 24.3 & 16.6 & 17.6 & 11.8 & 13.3
    & 33.4 & 36.0
    & 24.0 & 39.0\\
    XR-0
    & \textit{37.8} & 41.0 & \textit{24.5} & 27.1 & \textbf{17.1} & \textit{19.5} & \textbf{12.3} & \textit{14.7}
    & \textbf{34.6} & \textbf{36.9}
    & \textbf{26.9} & 40.1\\
    mXR-0
    & \textbf{39.0} & \textbf{42.8} & \textbf{24.9} & \textbf{28.6} & \textit{17.0} & \textbf{20.7} & \textit{12.1} & \textbf{15.7}
    & \textit{33.6} & \textit{36.8}
    & \textit{24.8} & \textit{42.2}\\
    \bottomrule
    \end{tabular}
    \caption{Report generation performance on the IU-XRay dataset. Scores are reported for both 10\% and 100\% training data.}
    \label{tab:repgen}
\end{table*}

\subsection{Fairness Analysis}
 
We perform bias analysis in order to assess model performance across demographic subgroups, guiding equitable and responsible deployment in clinical settings. By identifying disparities early, developers can implement targeted mitigation strategies to improve fairness and reliability.

We leverage the CheXpert dataset to evaluate sex- and age-related biases. Decoder models for disease classification are trained on specific subgroups and evaluated across others. For sex-specific analysis, models are trained on male, female, or all data, and tested on the male or female subset of the test set. Training on all data typically yields better performance due to increased data diversity. For each setting, we train a single model and apply bootstrapping with $n=200$ to report AUC scores.

\begin{figure}[!htbp]
    \centering
    \includegraphics{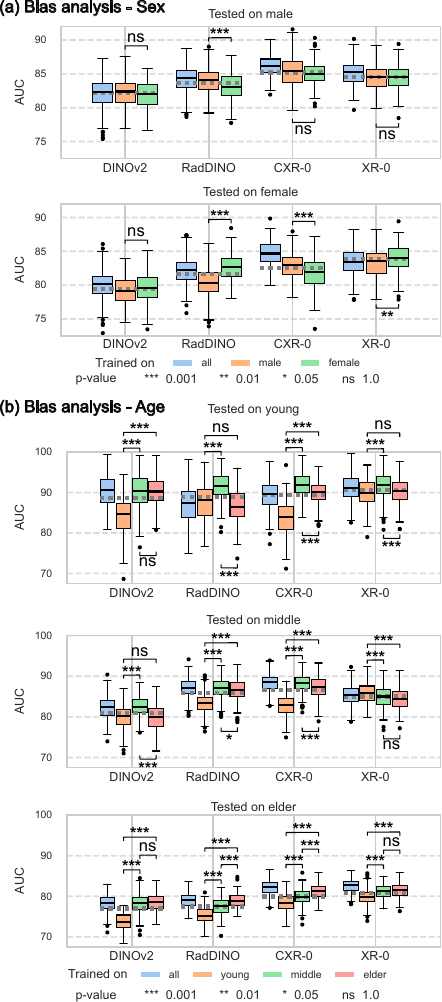}

    \caption{Fairness evaluation. \textbf{(a)} Results for sex-specific subgroups. \textbf{(b)} Fairness comparison across age group splits.}
    \label{fig:fairness}
    
\end{figure}

To assess statistical significance, we apply the Mann–Whitney test to the scores of models trained on different subgroups.
Models sharing the same backbone but trained on different subgroups are compared using this test. XR-0 and CXR-0 models show no significant differences when evaluated on male patients (Fig. \ref{fig:fairness}a). DINOv2 does not exhibit significant performance gaps across any subgroup, although it yields the lowest median score (indicated by dashed lines). Other configurations show significant differences, suggesting that DINOv2 is the least biased, followed by XR-0 and CXR-0. These findings underscore the importance of training on diverse datasets to improve performance across demographic subgroups.

We extend the analysis to age groups, where disease prevalence varies substantially. We define three age categories: young (0–35 years), middle-aged (35–60 years), and elderly (60+ years). We assess significance between each subgroup pairs (young-middle, young-elder, middle-elder). 
As shown in Fig. \ref{fig:fairness}b, shows XR-0, and DINOv2 exhibit the most non-significant performance gaps together (3 out of 9 comparisons). RadDINO follows with 1, and CXR-0 with none. These results indicate that while most models show some degree of bias depending on the subgroup, our models are among the least biased.
Notably, XR-0 and CXR-0 achieve the highest AUC scores for the elderly group, which is both the most prevalent and diagnostically complex due to subtle and overlapping findings.

A potential limitation of this analysis is the relatively small size of the test sets. While larger test sets may yield more robust conclusions, we expect the overall trends to remain consistent.

\section{Conclusion}
In this study, we introduced the first multi-anatomy X-ray foundation model, \textbf{XR-0}, and a chest-specific model \textbf{CXR-0}. All models were trained from random initialization to isolate the impact of pretraining data. Our dataset comprised approximately 1.2 million X-ray images spanning diverse anatomical regions. The models were evaluated across 12 datasets and a wide range of tasks, including classification, retrieval, segmentation, localization, visual grounding, and report generation.

Our work complements existing X-ray foundation models that primarily focus on chest anatomy. The results demonstrate that self-supervised learning enables models to learn generalizable features that transfer well to a wide-variety of clinical tasks. XR-0 achieves competitive, and in many cases state-of-the-art, performance—particularly on multi-anatomy and in-domain tasks—highlighting the value of diverse pretraining data. Chest-specialized models such as CXR-0 and RadDINO perform better on chest pathology detection and dense prediction tasks like segmentation, which suggests that anatomical focus remains important for certain applications. Notably, DINOv2 shows strong segmentation performance when paired with deep decoders like UPerNet, suggesting that large-scale natural image pretraining yields expressive low-level features well-suited for hierarchical decoding."

In early results, we also observe that the multimodal model mXR-0 excels in report generation, indicating that incorporating text supervision during pretraining can further enhance performance on generative tasks. These findings support the broader hypothesis that continual pretraining in a focused domain improves downstream task performance within that domain.

Foundation models are often expected to deliver superior performance, especially in zero-shot or few-shot scenarios. To test this, we conducted limited-data experiments on classification, report generation, and visual grounding. Our results show that high performance can be achieved using only 10\% of the training data, particularly when using specialized models like XR-0 and mXR-0. This has important implications for healthcare, where labeled data is often scarce and expensive to obtain.

Despite these promising results, our study has limitations. While we extensively evaluated the benefits of multi-anatomy pretraining, further experiments are needed to better understand the influence of training data composition and architectural choices. Future work will focus on collecting more multimodal data to improve performance on pathology detection tasks. We also plan to explore warm-start training, alternative hyperparameters (e.g., patch size, feature dimensions), and hierarchical backbones such as Swin Transformers~\citep{liu2021swin}, which may enhance performance on dense prediction tasks.

Finally, to ensure holistic evaluation and fairness, future studies should incorporate cross-dataset generalization tests and extend subgroup analyses to multi-anatomy datasets beyond chest imaging. This broader scope is essential for identifying potential biases and ensuring equitable model performance across diverse clinical contexts. We emphasize the need for non-chest X-ray collections that include sufficient sample sizes and rich metadata to support such evaluations.

In conclusion, we present a multi-anatomy X-ray foundation model and demonstrate its effectiveness across a broad spectrum of tasks. We believe this work lays the foundation for efficient, scalable, and generalizable AI solutions in radiology.

\bibliography{main}

\newpage
\appendix
\renewcommand{\thefigure}{A\arabic{figure}}
\renewcommand{\thetable}{A\arabic{table}}
\setcounter{figure}{0}
\setcounter{table}{0}

\begin{appendices}

\section{List of Supported Classes in the Quality Control Dataset}
\label{apdx:qc_classes}

\begin{table}[h!]
\begin{tabular}{lcp{4cm}}
\toprule
\textbf{Task} & \textbf{Num Classes} & \textbf{Classes} \\
\midrule
Anatomy Classification          & 13 & chest, abdomen, pelvis, shoulder, hip, hand, knee, foot, wrist, babygram, ankle, elbow, others \\
\hline
View Classification       & 6 & 1. FRONTAL \newline 2. LATERAL, LATERALL, LATERALR\newline 3. PA BILAT, AP BILAT, FRONTAL BILAT \newline 4. SUN BILAT, SCAPHOID \newline 5. BABYGRAM, LATERAL BILAT\newline 6. OTHERS \\
\hline
Mirror Classification  & 5 & 1. LATERALL\newline 2. LATERALR\newline 3. XTLLL\newline 4. XTLLR\newline 5. FRONTAL, LATERAL, FRONTAL BILAT, LATERAL BILAT, PA BILAT, AP BILAT, SUN BILAT, SCAPHOID, BABYGRAM, OTHERS \\
\hline
Lead Marker Classification & 2 & With and Without Lead Marker \\
\bottomrule
\end{tabular}
\end{table}
\newpage

\section{Multimodal Pretraining}
\label{apdx:multimodal_pretraining}
To extend XR-0 into a multimodal model (mXR-0), we incorporate paired clinical reports and add a text encoder. For multimodal pretraining, we initialize the vision encoder with the pretrained XR-0 weights and adopt PubMedBERT~\citep{pubmedbert}, a language model pretrained on biomedical literature, as the text encoder. Both encoders are jointly fine-tuned using an image–text contrastive learning objective inspired by CLIP~\citep{clip}, which aligns image and text embeddings in a shared multimodal space by pulling together matched pairs and pushing apart mismatched ones. Both encoders output 768-dimensional representations, and we apply a linear projection to the \texttt{[CLS]} token of each encoder to map their outputs into a shared 512-dimensional multimodal embedding space. Similar to the training of XR-0, a cosine learning rate scheduler with a linear warmup during the first 10 epochs was used, paired with a weight decay schedule ranging from 0.04 to 0.2. Different peak learning rates were assigned to different components: $1e^{-4}$ for the linear projection heads, and $1e^{-5}$ for both the pretrained vision and text encoders. This setup promotes effective adaptation to text supervision while preserving the pretrained knowledge acquired in the pretrained XR-0. We trained the multimodal mXR-0 for 100 epochs with a batch size of 80 per GPU on a 8 NVIDIA A10G GPUs. We select the best checkpoint based on performance in the CheXpert image–text retrieval evaluation task described earlier.

For multimodal pretraining, we use a subset of the training data containing approximately 69,000 image–report pairs, with each report associated with an average of 2.4 images. We reserve 10\% of this dataset for testing, resulting in 62,000 pairs used for training. To enhance the quality of contrastive learning, we design a report preprocessing pipeline that cleans, extracts, and structures relevant content from raw medical text, focusing on radiology-specific sections such as "Findings" or "Report" (See Appendix \ref{apdx:processed_reports}). The preprocessing includes:
(1) extracting the main content by locating keywords like "FINDINGS" or "REPORT" and removing irrelevant preamble text;
(2) cleaning and segmenting the extracted content into meaningful sentences, while removing noise such as technical metadata and formatting artifacts;
(3) filtering out short or incomplete sentences and merging them where appropriate to preserve context and improve readability;
(4) generating a shorter version of the report by combining every two sentences, resulting in a list of condensed sentences per report. During training, a preprocessed short report and an image are randomly sampled from the same report–image pair to form a positive input pair for contrastive learning. Image processing and augmentations follow the same settings used for the global crops in XR-0 training, including random cropping (50\%–100\%), horizontal flipping, rotation, auto-contrast, histogram equalization, and Gaussian blur.

\subsection{Processed Report Examples}
\label{apdx:processed_reports}
\begin{figure}[h]
    \centering
    \includegraphics[width=\textwidth]{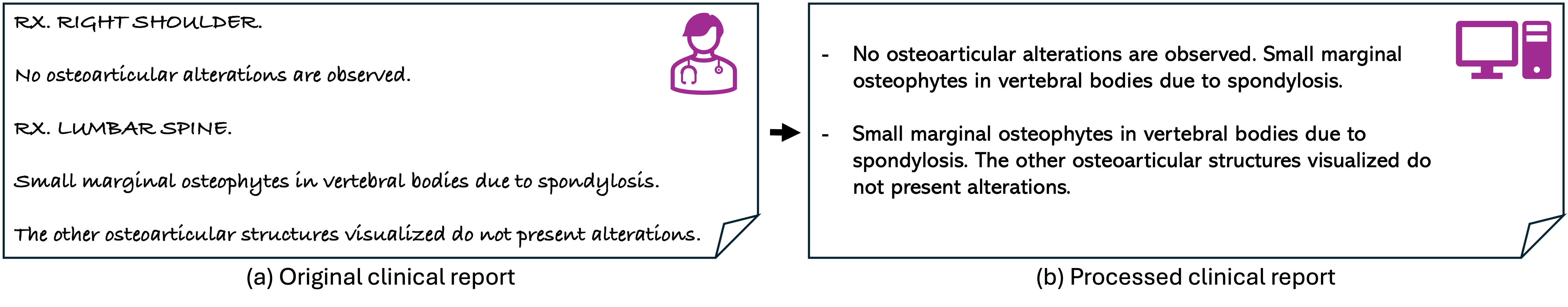}
    \caption{An example of an original unprocessed report and the final processed output used for the pretraining of multimodal model mXR-0.}
    \label{fig:processed_reports}
\end{figure}

\newpage

\section{Additional Image Retrieval Examples}
\label{apdx:retrieval_examples}
\begin{figure}[h]
    \centering
    \includegraphics[height=20cm]{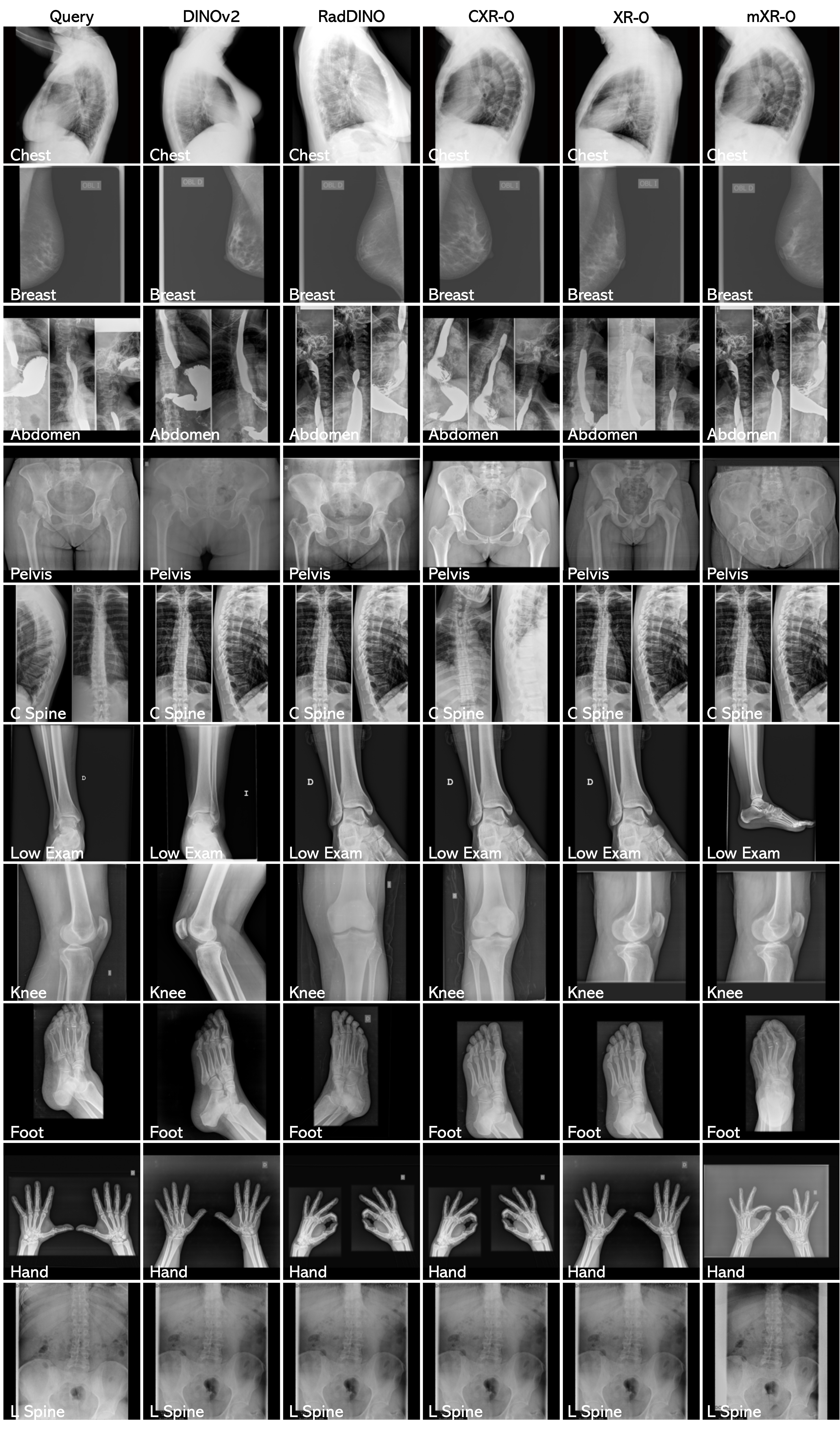}
    \caption{Additional qualitative results for image retrieval.}
    \label{fig:retrieval_examples}
\end{figure}

\newpage

\section{PCA Visualization of the Patch Embeddings}
\label{apdx:pca_of_patchemb}

Visualizing the patch embeddings with PCA indicates that the spatial features contain information related to the land markers. A classifier on the patch embeddings can better detect the land markers compared to a [CLS] token-based model.

\begin{figure}[h]
    \centering
    \begin{minipage}{0.6\textwidth}
        \centering
        \includegraphics[width=\linewidth]{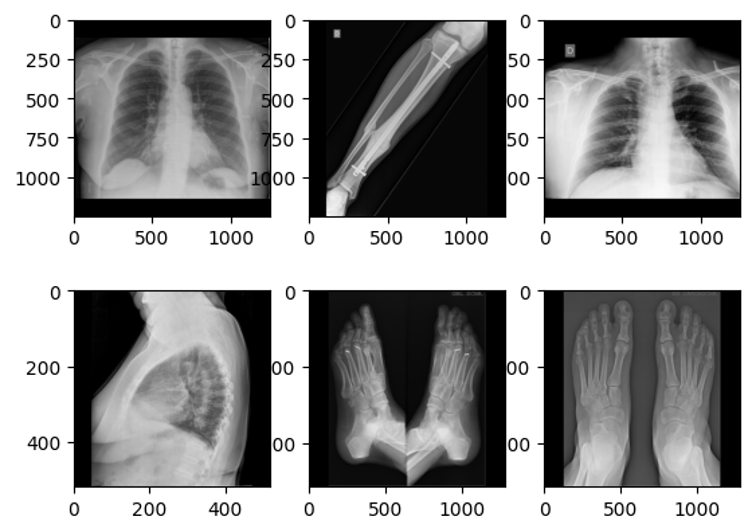}
        \caption{Example multi-anatomy images with 2 images containing lead marker (top row middle and right items).}
    \end{minipage}
    
    \begin{minipage}{0.6\textwidth}
        \centering
        \includegraphics[width=\linewidth]{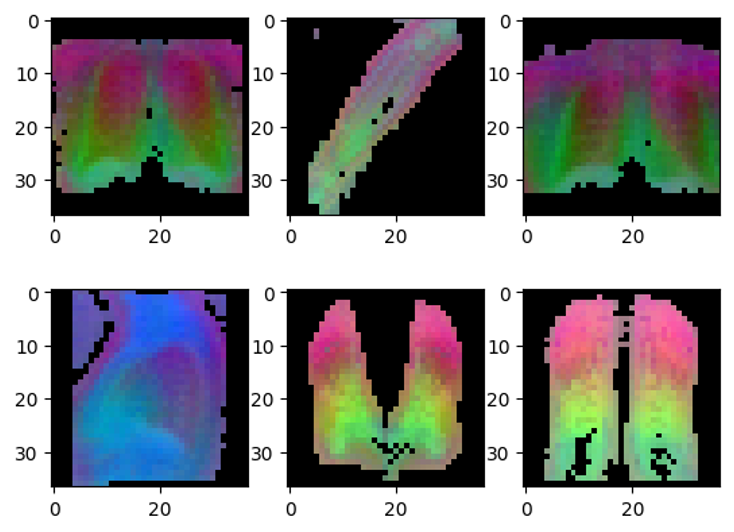}
        \caption{PCA visualization of the patch embeddings.}
    \end{minipage}
\end{figure}

\newpage

\section{Localization Examples}
\label{apdx:localization_examples}
\begin{figure}[h]
    \centering
    \includegraphics[height=20cm]{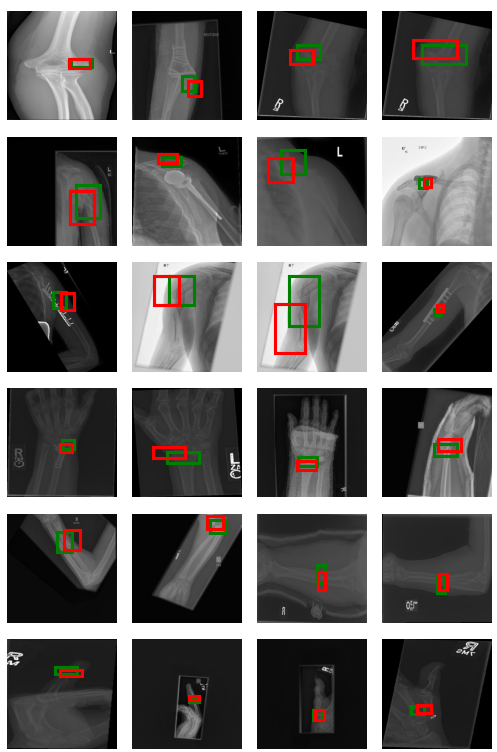}
    \caption{Localization qualitative results.}
    \label{fig:localization_examples}
\end{figure}

\newpage

\section{Visual Grounding Examples}
\label{apdx:grounding_examples}
\begin{figure}[h]
    \centering
    \includegraphics[width=0.6\textwidth]{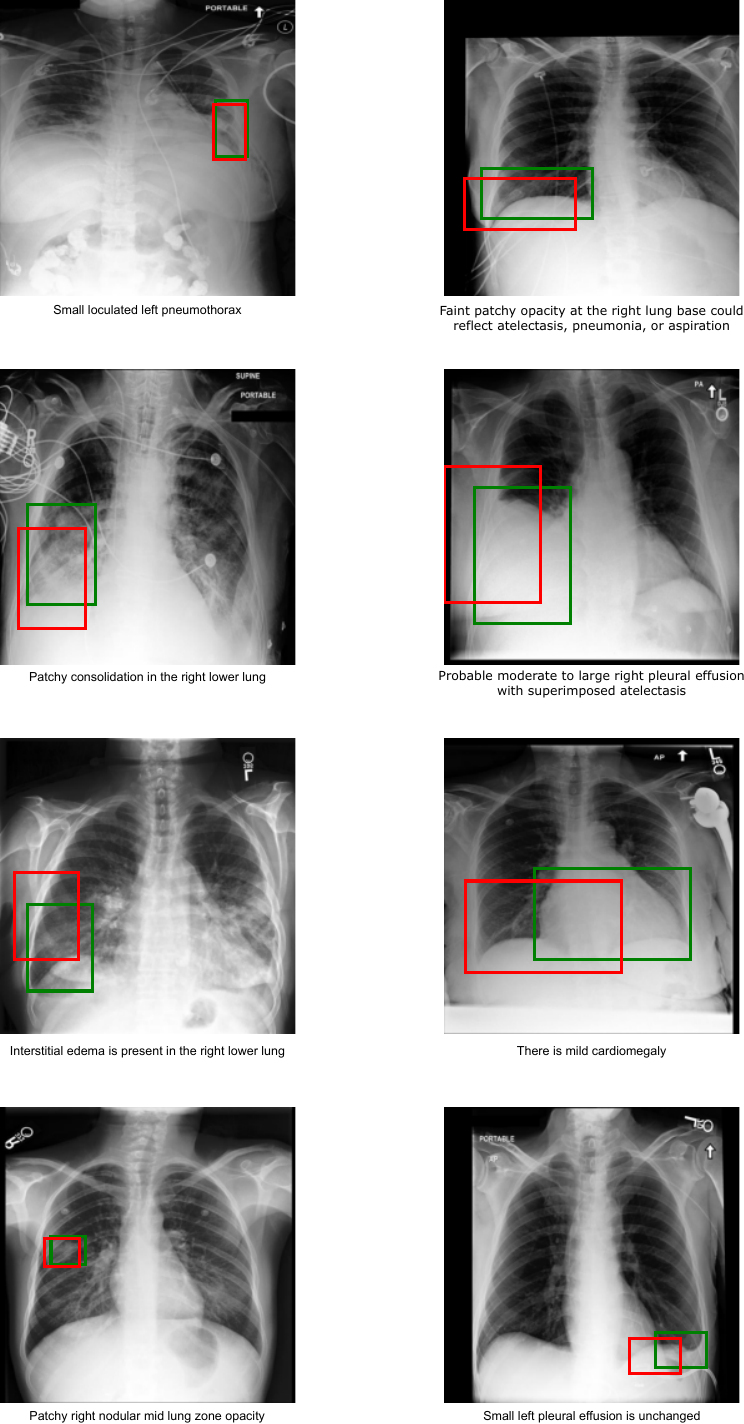}
    \caption{Visual grounding qualitative results.}
    \label{fig:grounding_examples}
\end{figure}
\newpage

\section{Report Generation Examples}
\label{apdx:repgen_examples}
\begin{figure}[h]
    \includegraphics[width=\linewidth]{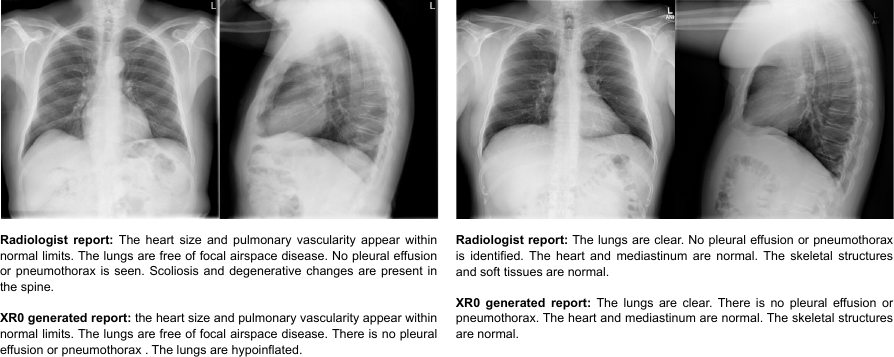}
    \caption{Report generation qualitative results.}
    \label{fig:repgen}
\end{figure}
\newpage

\newpage

\end{appendices}

\end{document}